%% file: acl_latex.tex
\pdfoutput=1
\documentclass[11pt]{article}

\usepackage{acl}
\usepackage{booktabs}
\usepackage{multirow}
\usepackage{colortbl}
\usepackage{enumitem}
\usepackage{xcolor}

\definecolor{highlight}{RGB}{255, 255, 200}

\usepackage{times}
\usepackage{latexsym}
\usepackage{multirow}
\usepackage{booktabs}
\usepackage{xltabular}
\usepackage{makecell}
\usepackage{filecontents}
\usepackage[T1]{fontenc}
\usepackage{amsmath}
\usepackage[utf8]{inputenc}

\usepackage{microtype}

\usepackage{inconsolata}

\usepackage{graphicx}

\title{DocCHA: Towards LLM-Augmented Interactive Online diagnosis System}

\author{
Xinyi Liu$^{1}$, Dachun Sun$^{1}$, Yi R. Fung$^{2}$, Dilek Hakkani-Tür$^{1}$, Tarek Abdelzaher$^{1}$  \\
$^{1}$University of Illinois Urbana-Champaign, USA \\
$^{2}$The Hong Kong University of Science and Technology, Hong Kong SAR, China \\
\texttt{\{liu323, dsun18, dilek, zaher\}@illinois.edu}, \texttt{yrfung@ust.hk}
}

\begin{document}
\maketitle
\begin{abstract}
Despite the impressive capabilities of Large Language Models (LLMs), existing Conversational Health Agents (CHAs) remain static and brittle, incapable of adaptive multi-turn reasoning, symptom clarification, or transparent decision-making. This hinders their real-world applicability in clinical diagnosis, where iterative and structured dialogue is essential. We propose \textbf{DocCHA}, a confidence-aware, modular framework that emulates clinical reasoning by decomposing the diagnostic process into three stages: (1) symptom elicitation, (2) history acquisition, and (3) causal graph construction. Each module uses interpretable confidence scores to guide adaptive questioning, prioritize informative clarifications, and refine weak reasoning links.

Evaluated on two real-world Chinese consultation datasets (IMCS21, DX), DocCHA consistently outperforms strong prompting-based LLM baselines (GPT-3.5, GPT-4o, LLaMA-3), achieving up to 5.18\% higher diagnostic accuracy and over 30\% improvement in symptom recall, with only modest increase in dialogue turns. These results demonstrate DocCHA’s effectiveness in enabling structured, transparent, and efficient diagnostic conversations—paving the way for trustworthy LLM-powered clinical assistants in multilingual and resource-constrained settings.
\end{abstract}

\section{Introduction}
\begin{figure*}[ht]
\centering
\includegraphics[width = 1\textwidth]{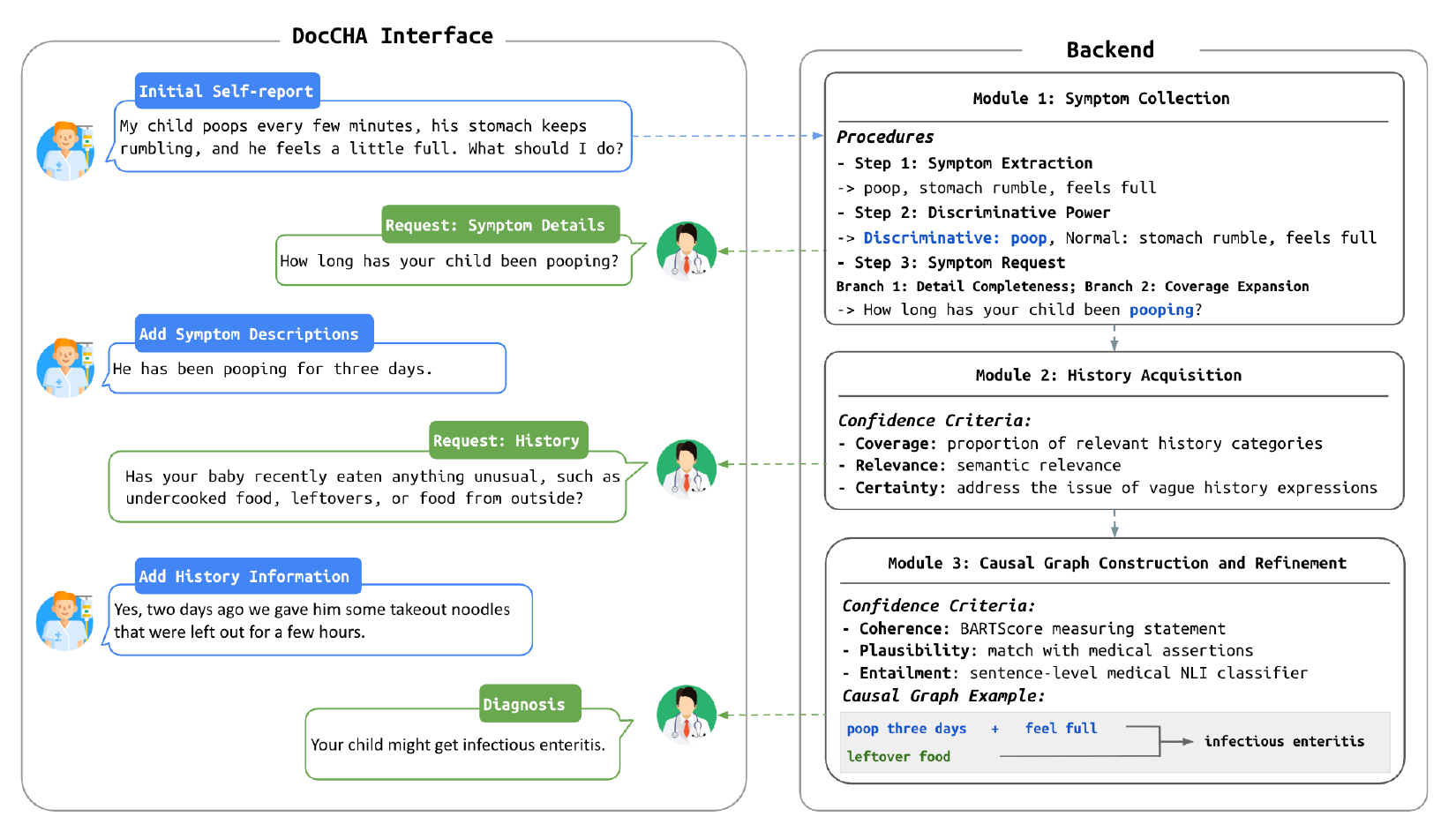}
\caption{Overview of the DocCHA framework. The left panel shows a multi-turn dialogue between the user and the diagnostic agent, while the right panel illustrates the corresponding backend modules, including symptom collection, history acquisition, and causal reasoning.}
\label{fig:framework}
\end{figure*}

Large Language Models (LLMs) have dramatically enhanced the capabilities of conversational agents across domains such as education \cite{mctear2022conversational}, climate policy \cite{vaghefi2023chatclimate}, and healthcare \cite{montenegro2019survey}. In the medical domain, LLM-based Conversational Health Agents (CHAs) have been deployed for tasks including health report generation \cite{liu2023biosignal}, patient education \cite{abbasian2023conversational}, and clinical information extraction \cite{chen2023utility}. Existing approaches typically fall into three paradigms: (i) zero-shot or few-shot prompting of general-purpose LLMs \cite{biswas2023chatgpt}, (ii) fine-tuned domain-specific models \cite{li2023chatdoctor, han2023medalpaca}, and (iii) multimodal extensions incorporating imaging or physiological signals \cite{xu2023elixr, belyaeva2023multimodal}.

Despite recent advances, most current CHAs lack the core reasoning behaviors expected in real-world clinical workflows. They rely on static prompts or scripted flows that fail to account for ambiguous, incomplete, or evolving user input. In contrast, clinical diagnosis is a highly iterative process—clinicians adaptively identify missing information, elicit symptom nuances (e.g., severity, progression), integrate patient history, and reason causally across heterogeneous data sources \cite{elstein2002clinical}. These dynamic and contingent behaviors are not well captured by existing LLM-based CHAs, leading to oversimplified questioning and unreliable diagnostic decisions.

To address this gap, we introduce \textbf{DocCHA}, a modular, confidence-aware framework designed to emulate human-like diagnostic reasoning in multi-turn dialogue settings. As shown in Figure~\ref{fig:framework}, DocCHA decomposes the diagnostic process into three core stages—symptom elicitation, history acquisition, and causal reasoning—each powered by dedicated modules that operate under confidence- and quota-based control. This modularity enables DocCHA to adaptively probe information gaps, prioritize high-yield clarifications, and construct interpretable causal graphs that justify diagnostic conclusions.

\begin{itemize}[leftmargin=*,itemsep=0pt,topsep=0pt]
    \item \textbf{Symptom Collection Module} identifies high-impact symptoms based on discriminative diagnostic value and completeness across structured attributes (e.g., duration, severity). A dual-score controller (coverage + detail) determines whether to ask about new symptoms or refine existing ones.
    \item \textbf{History Acquisition Module} targets contextual background (e.g., exposure, medication use, chronic disease), scoring coverage, semantic relevance to diagnosis candidates, and expression certainty to control questioning depth.
    \item \textbf{Causal Graph Construction Module} integrates collected symptoms and history into a structured diagnostic reasoning chain. The graph is scored on clinical coherence and alignment with medical knowledge bases (e.g., UMLS), and weak links trigger targeted follow-up.
\end{itemize}

\begin{figure*}[ht]
    \centering
    \includegraphics[width = 1\textwidth]{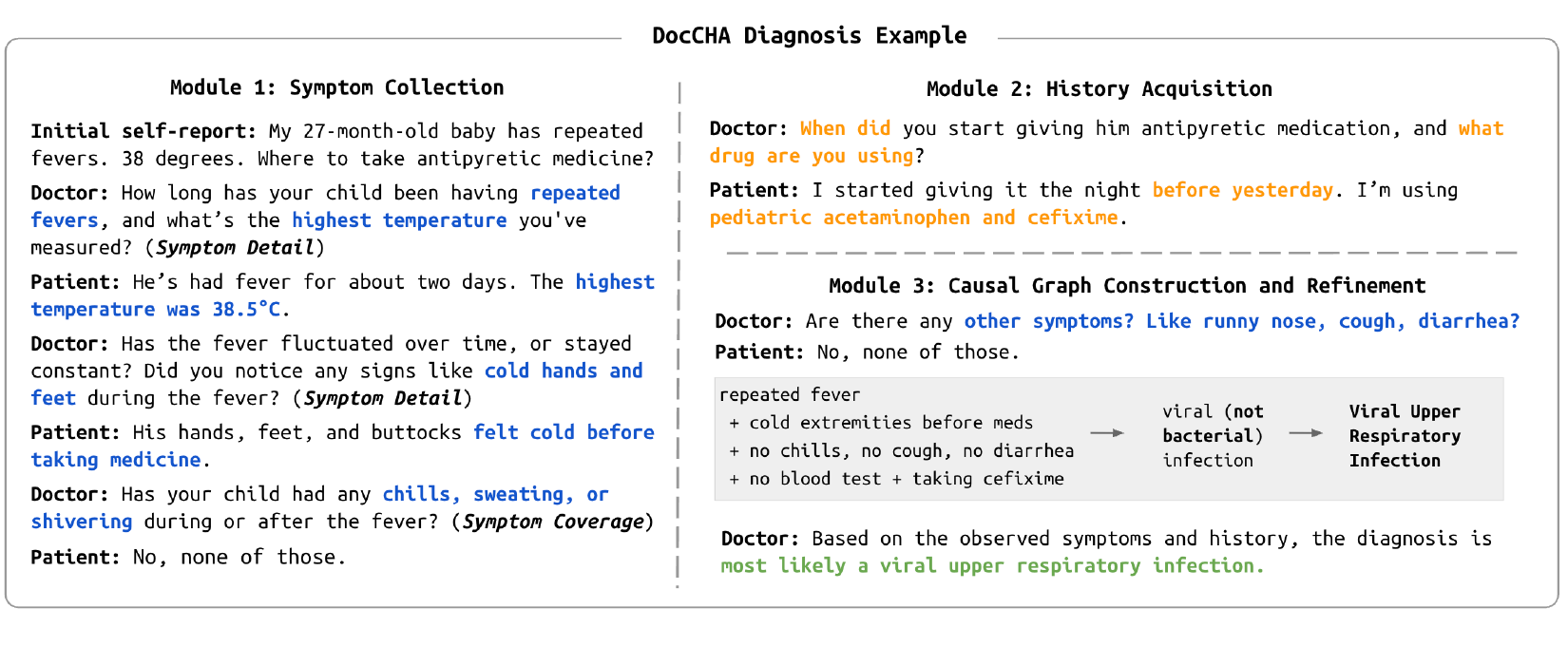}
\caption{An end-to-end diagnosis example with DocCHA. The left shows the multi-turn interaction between patient and system; the right illustrates the corresponding backend modules that drive symptom elicitation, history acquisition, and causal reasoning.}

    \label{fig:diagnosis_llm}
\end{figure*}

Each module operates autonomously yet communicates through a shared confidence state, resulting in an \textbf{adaptive, interpretable, and budget-aware} dialogue system. A full interaction example is illustrated in Figure~\ref{fig:diagnosis_llm}, showing how DocCHA dynamically shifts focus across modules to build an accurate and explainable diagnosis.

Extensive experiments on two Chinese diagnostic dialogue datasets—IMCS21 and DX—demonstrate that DocCHA significantly improves performance over prompting-only baselines. Specifically, it achieves up to \textbf{+5.18\%} improvement in accuracy over GPT-4o and up to \textbf{+29.63\%} over LLaMA-3, while extracting over \textbf{30\%} more medically relevant information under the same turn budget. These results validate that structure-aware, confidence-guided reasoning offers distinct advantages beyond LLM scale.

\vspace{0.5em}
\noindent \textbf{Our contributions are summarized as follows:}
\begin{itemize}[leftmargin=*,itemsep=0pt,topsep=0pt]
    \item We propose \textbf{DocCHA}, a modular and confidence-driven diagnostic framework that emulates multi-turn clinical reasoning through structured symptom elicitation, history probing, and causal graph refinement.
    \item We develop novel scoring mechanisms for symptom detail utility, history confidence estimation, and weak-link identification in causal reasoning.
    \item We conduct thorough evaluations showing that DocCHA significantly outperforms state-of-the-art prompting baselines in both accuracy and informativeness, while preserving interaction efficiency.
\end{itemize}

\section{Related Work}
\paragraph{Conversational Health Agents with LLMs}

Recent work on LLM-based Conversational Health Agents (CHAs) falls into three categories: general-purpose prompting, domain-specific fine-tuning, and multimodal systems. General-purpose CHAs rely on off-the-shelf LLMs to answer medical questions or synthesize content, such as generating treatment plans \cite{chen2023utility}, summarizing biosignals \cite{liu2023biosignal}, or detecting patient stress \cite{abbasian2023conversational}. Domain-adapted models like ChatDoctor \cite{li2023chatdoctor}, MedAlpaca \cite{han2023medalpaca}, and BioGPT \cite{luo2022biogpt} improve performance by training on medical corpora. Multimodal approaches, including ELIXR \cite{xu2023elixr} and HeLM \cite{belyaeva2023multimodal}, integrate clinical images or structured inputs with LLM reasoning.

Despite their strengths, most CHAs remain static—incapable of adaptive multi-turn interactions or structured diagnostic reasoning. They often lack mechanisms to assess information sufficiency, handle ambiguity, or explain their conclusions. In contrast, DocCHA is designed to be modular and adaptive: it conducts interactive diagnosis through confidence-scored symptom and history collection, and produces interpretable causal graphs—without requiring additional fine-tuning or multimodal input.

\paragraph{Information Sufficiency, Uncertainty Modeling, and Causal Reasoning}

Effective clinical dialogue requires not only recognizing symptoms, but also collecting detailed attributes and relevant history. Prior systems like MedDialog \cite{yang2023meddialog} and CareChat \cite{lin2023carechat} simulate interactions but do not explicitly model information completeness. DocCHA addresses this by assigning confidence scores to each stage—capturing symptom coverage and detail, history relevance and certainty—to guide budget-aware, adaptive questioning.

Uncertainty modeling has been explored in tasks like claim verification or vague statement detection \cite{zhang2022bioie, honovich2022true}; DocCHA extends these ideas to drive clarification within dialogue, identifying speculative or underspecified responses.

Finally, while diagnosis is inherently causal, existing CHAs often lack structured reasoning paths. DocCHA builds explicit causal graphs linking symptoms and history to diagnoses, scored by coherence, entailment, and medical plausibility. It validates these graphs against UMLS \cite{bodenreider2004umls} and SemMedDB \cite{kilicoglu2012semmeddb}, enabling interpretable, evidence-grounded reasoning.

\section{Method}

We propose \textbf{DocCHA}, a modular, confidence-driven diagnostic framework that emulates multi-turn clinical reasoning with LLMs. The architecture consists of three sequential modules: 
(1) \textit{Symptom Collection}, 
(2) \textit{History Acquisition}, and 
(3) \textit{Causal Graph Construction and Refinement}. 

Each module integrates a dedicated confidence function to evaluate information sufficiency and trigger follow-up questions. These metrics are explicitly designed to maximize diagnostic discriminability \cite{zhou2019learning}, reduce uncertainty \cite{zhang2022bioie}, and support structured causal reasoning \cite{yuan2021bartscore, bodenreider2004umls}. The prompts are shown in Session \ref{sec:appendix}

To support downstream analysis and evaluation, we additionally introduce a \textit{Patient Agent Simulator}, which simulates grounded patient responses from real-world consultations.

\subsection{Module 1: Symptom Collection}

Effective diagnosis requires not only knowing what symptoms are present but also how precisely they are described. In real-world consultations, patients frequently omit symptoms they deem irrelevant or fail to specify important attributes such as onset time, severity, or duration. These omissions significantly reduce diagnostic accuracy. A naïve approach is to let LLMs directly generate follow-up questions based on the current conversation context. However, such free-form prompting often leads to inefficient, redundant, or unfocused questioning—e.g., repeatedly asking about common symptoms regardless of their diagnostic utility. 

To address this, we introduce a structured and confidence-guided framework that formalizes symptom collection as a dynamic optimization process, balancing informativeness, specificity, and interaction cost. Our design is rooted in a key diagnostic insight: not all symptoms are equally valuable, and eliciting additional detail should be prioritized for symptoms that meaningfully differentiate between likely diagnoses. Following prior work on feature discriminability in clinical prediction tasks \cite{zhou2019learning}, we quantify each symptom’s discriminative power to guide targeted questioning. In contrast to free prompting, this strategy ensures that every follow-up maximally reduces diagnostic uncertainty under a fixed query budget.

Let $S_u = \{s_1, s_2, \dots, s_n\}$ denote the set of symptoms extracted from the patient’s utterances. We first use an LLM to obtain the top-$K$ candidate diagnoses $\mathcal{D} = \{d_1, \dots, d_K\}$. For each $s_i \in S_u$, we compute its \textbf{Discriminative Power} $DP(s_i)$ as the variance of its conditional probabilities across the candidate diagnoses:

\begin{equation}
DP(s_i) = \operatorname{Var}_{d_k \in \mathcal{D}} P(s_i | d_k)
\end{equation}

This scores how informative $s_i$ is for distinguishing between hypotheses. To assess the overall quality of the extracted symptoms, we decompose the module confidence into two dimensions: coverage and detailness.

The \textbf{Symptom Coverage Confidence} $C_{\text{cov}}$ reflects the extent to which the known symptoms match the canonical symptom set of the most likely diagnosis:

\begin{equation}
C_{\text{cov}} = \max_{d_k \in \mathcal{D}} \frac{|S_u \cap S_{d_k}|}{|S_{d_k}|}
\end{equation}

The \textbf{Symptom Detail Confidence} $C_{\text{det}}$ evaluates how many required attributes (e.g., onset, frequency, severity) have been filled in for each reported symptom:

\begin{equation}
C_{\text{det}} = \frac{1}{|S_u|} \sum_{s_i \in S_u} \frac{|A(s_i)|}{|A_{\text{req}}(s_i)|}
\end{equation}

The overall confidence for the symptom collection stage is a weighted combination:

\begin{equation}
C_{\text{sym}} = \alpha C_{\text{cov}} + (1 - \alpha) C_{\text{det}}
\end{equation}

If $C_{\text{sym}}$ falls below a threshold $\tau_{\text{sym}}$ and the number of follow-up questions is below a quota $q_{\text{sym}}$, DocCHA generates targeted clarification questions. These are directed toward symptoms with high $DP(s_i)$ to maximize diagnostic value per query.

This design offers two key advantages over open-ended prompting. First, it introduces explicit diagnostic reasoning into the questioning loop, ensuring alignment with clinical objectives rather than relying on heuristic LLM behavior. Second, it provides modular interpretability: low coverage versus low detailness lead to qualitatively different follow-up strategies (e.g., exploring missing symptoms versus probing known ones). Overall, this structured framework transforms symptom elicitation from a passive task into an active, explainable diagnostic optimization process.

\subsection{Module 2: History Acquisition}

Contextual factors—such as travel, exposure, medication use, and chronic conditions—often determine the correct diagnosis among competing candidates. Unlike symptoms, these details are rarely volunteered unprompted, and generic prompts like ``Do you have any medical history?'' are too vague. Prior work highlights both the importance of structured history elicitation~\cite{peng2023selfaligning, lin2023carechat} and the risks of vague or unsupported inputs~\cite{zhang2022bioie}.

DocCHA reframes history acquisition as a confidence-scored, goal-directed process that dynamically prioritizes the most diagnostically relevant aspects of patient background.

Let $H_u = \{h_1, \dots, h_m\}$ denote extracted history items from user input, and let $H_{\mathcal{D}}$ be the union of expected history categories associated with the top-$K$ candidate diagnoses $\mathcal{D} = \{d_1, \dots, d_K\}$. We compute a history sufficiency score using three complementary dimensions:

\begin{itemize}[leftmargin=*, itemsep=0pt]
    \item \textbf{Coverage ($C_{\text{cov}}$)}: the proportion of relevant categories addressed:
    \begin{equation}
    C_{\text{cov}} = \frac{|H_u \cap H_{\mathcal{D}}|}{|H_{\mathcal{D}}|}
    \end{equation}

    \item \textbf{Relevance ($C_{\text{rel}}$)}: the semantic similarity between $h_i$ and the patient's overall context ($S_u \cup \mathcal{D}$), using sentence embeddings:
    \begin{equation}
    C_{\text{rel}} = \frac{1}{|H_u|} \sum_{h_i \in H_u} \cos(\mathbf{v}_{h_i}, \mathbf{v}_{S_u \cup \mathcal{D}})
    \end{equation}

    \item \textbf{Certainty ($C_{\text{cert}}$)}: the average confidence of history expressions, scored by an LLM-based classifier:
    \begin{equation}
    C_{\text{cert}} = \frac{1}{|H_u|} \sum_{h_i \in H_u} \operatorname{LLMScore}(h_i)
    \end{equation}
\end{itemize}

The final sufficiency score is a weighted combination:
\begin{equation}
C_{\text{hist}} = \lambda_1 C_{\text{cov}} + \lambda_2 C_{\text{rel}} + \lambda_3 C_{\text{cert}}
\end{equation}

When $C_{\text{hist}} < \tau_{\text{hist}}$, DocCHA triggers targeted follow-up questions—focusing on missing categories, low-relevance entries, or vague expressions. Compared to open-ended prompting, this enables more efficient and clinically aligned history elicitation. Moreover, the decomposed scores inform downstream modules (e.g., causal reasoning) of which dimensions are well-grounded versus uncertain, enhancing overall interpretability and robustness.

\subsection{Module 3: Causal Graph Construction and Refinement}

To emulate the reasoning patterns of trained clinicians, DocCHA constructs an explicit causal graph that links symptoms and historical factors to a likely diagnosis. This step is not only important for providing interpretable outputs, but also for enabling traceable diagnostic logic—something lacking in most LLM-based systems, which often generate flat, opaque diagnosis results. While recent studies show that LLMs can produce plausible medical justifications when prompted, such explanations often suffer from hallucinated connections or circular reasoning. Moreover, without a structured representation, it becomes difficult to quantify which parts of the reasoning chain are weak or speculative.

To address these limitations, we frame the diagnostic reasoning process as causal graph construction. Let $G = (V, E)$ be a directed acyclic graph, where nodes $V$ correspond to observed symptoms, historical findings, and candidate diagnoses, and directed edges $E$ denote inferred causal or temporal relations. Each edge $e = (a \rightarrow b)$ reflects an LLM-generated claim that $a$ contributes to or precedes $b$ in a medically meaningful way. However, since LLMs may generate incorrect or shallow relations, we do not treat the graph as immutable. Instead, we introduce a multi-factor scoring function that assesses the overall plausibility and completeness of $G$, and triggers follow-up clarification questions when the confidence is low.

The final graph confidence $C_{\text{causal}}$ integrates three components: coherence, plausibility, and entailment. Coherence is measured using a language-model-based scorer (e.g., BARTScore) over the sequence of reasoning statements. Medical plausibility is estimated by checking whether the inferred edges match known medical assertions from curated resources such as UMLS and SemMedDB:

\begin{equation}
C_{\text{med}} = \frac{|E \cap E_{\text{UMLS}}|}{|E|}
\end{equation}

In parallel, we compute edge-wise entailment probabilities using a sentence-level medical NLI classifier. The final confidence score aggregates these components:

\begin{equation}
C_{\text{causal}} = \mu_1 C_{\text{coh}} + \mu_2 C_{\text{med}} + \mu_3 C_{\text{entail}}
\end{equation}

If $C_{\text{causal}}$ falls below a threshold $\tau_{\text{causal}}$, DocCHA identifies the weakest edge—typically the one with low entailment or unsupported by medical databases—and generates a targeted clarification prompt, such as asking for temporal order, epidemiological linkage, or missing conditions.

This design provides two distinct advantages over prompting-based justification methods. First, it enables a quantitative and modular evaluation of reasoning quality, where each causal link can be inspected, validated, or revised independently. Second, by leveraging external knowledge graphs and internal confidence signals, we prevent the system from prematurely committing to speculative diagnoses. Ultimately, this approach brings structure and accountability to LLM-based diagnostic reasoning, bridging the gap between generative fluency and clinical rigor.

\subsection{Patient Agent Simulator (for Evaluation)}

Evaluating DocCHA requires more than final-diagnosis accuracy—it also demands a principled way to assess how effectively the system elicits clinically relevant information. Manual annotation for each exchange is costly and subjective, so we introduce a \textit{Patient Agent Simulator} for automated, reproducible evaluation of information-seeking quality.

The simulator uses real consultation transcripts $T$ as ground truth. When queried with a follow-up question $q_i$, it returns a response $a_i$ based on whether the answer is grounded in $T$:

\begin{equation}
a_i = 
\begin{cases}
\texttt{LLM}(T, q_i), & \text{if answerable from } T \\
\text{``I didn’t notice that.''}, & \text{otherwise}
\end{cases}
\end{equation}

This method adapts techniques from controlled summarization and factual consistency~\cite{honovich2022true}, ensuring patient responses remain grounded in real-world data rather than hallucinated dialogue.

We define \textit{information extraction coverage} as the proportion of medically relevant facts in $T$ retrieved by DocCHA’s questions. Let $I_T$ be all extractable facts in $T$, and $I_{\text{CHA}}$ the set elicited by DocCHA:

\begin{equation}
\text{Coverage} = \frac{|I_T \cap I_{\text{CHA}}|}{|I_T|}
\end{equation}

This simulation-based setup allows precise, scalable, and interpretable evaluation of DocCHA’s information elicitation capabilities, enabling fair comparison across models and prompting strategies. Notably, the Patient Agent is used only during evaluation, isolating the quality of interaction without influencing system behavior.

\section{Experiments}
\label{exp}

We evaluate DocCHA on two key capabilities essential for interactive medical diagnosis: (1) accurately predicting final diagnoses through multi-turn patient interaction, and (2) eliciting clinically critical information with the coverage and precision of real doctors. Experiments are conducted on two real-world doctor-patient dialogue datasets, using matched interaction budgets and comparisons with strong prompting baselines.
\input{tables/evaluation_res}
\subsection{Datasets and Setup}
\label{sec:dataset}

\noindent\textbf{IMCS21.}  
This dataset contains real pediatric online consultations in Chinese, covering ten common diagnoses such as neonatal jaundice and bronchitis. Each case starts with a free-form self-report followed by multiple doctor-initiated questions. We use the Google Translate API to convert samples into English, and select 100 sessions for evaluation.

\noindent\textbf{DX Dataset.}  
The DX dataset is a Chinese open-domain medical QA benchmark. Each sample contains a brief patient description, gold diagnosis, and structured multi-turn follow-up. We sample 120 cases with at least three rounds of questioning. Translations are aligned to UMLS categories to ensure cross-model consistency.

\subsection{Experimental Setting}

We evaluate DocCHA using GPT-4o\footnote{\url{https://openai.com/index/gpt-4o}} and LLaMA-3\footnote{\url{https://github.com/meta-llama/llama3}} as backbones, representing proprietary and open-source models. All agents interact with the same Patient Agent in a multi-turn dialogue, capped by matched interaction budgets.

To isolate the effect of DocCHA’s structured pipeline, we compare against:

\begin{itemize}[leftmargin=*,itemsep=0pt,topsep=0pt]
  \item \textbf{LLaMA-3 Direct Prompting}: Identical interface, no modular pipeline.
  \item \textbf{GPT-3.5 Direct Prompting}: A lower-capacity variant with same prompting structure.
  \item \textbf{GPT-4o Direct Prompting}: GPT-4o engages freely with the patient, but without confidence-driven modules. Turn count is matched to DocCHA.
\end{itemize}

All baselines ask the same number of questions and receive identical Patient Agent responses, ensuring a fair comparison focused on pipeline design.

\paragraph{Weight Settings.}  
Module weights are fixed based on clinical reasoning principles:
\begin{itemize}[leftmargin=*,itemsep=0pt,topsep=0pt]
  \item \textbf{Symptom Module}: $\alpha = 0.5$ balances coverage and attribute detail (onset, severity), supporting ambiguity reduction.
  \item \textbf{History Module}: $\lambda_1 = 0.4$ (coverage), $\lambda_2 = 0.3$ (relevance), $\lambda_3 = 0.3$ (certainty). Prioritizes broad screening before fluency evaluation.
  \item \textbf{Causal Graph Module}: $\mu_1 = 0.4$ (coherence), $\mu_2 = 0.3$ (UMLS grounding), $\mu_3 = 0.3$ (entailment). Emphasizes logical integrity in diagnostic chains.
\end{itemize}

Each module is allocated up to 4 questions. Confidence scores are used internally for decision-making. Similarity and entailment use \texttt{mpnet-base-v2} \cite{song2020mpnet} and a domain-adapted MedNLI model. Factual grounding references UMLS \cite{bodenreider2004umls} and SemMedDB\footnote{\url{https://semmeddb.nlm.nih.gov/}}.

This controlled setup enables precise attribution of diagnostic gains to DocCHA’s modular, confidence-driven design.

\subsection{Evaluation Metrics}

To capture both diagnostic accuracy and information-seeking behavior, we report on:
\begin{itemize}[leftmargin=*,itemsep=0pt,topsep=0pt]
    \item \textbf{Accuracy (Acc.)}: Exact match between predicted and gold diagnosis.
    \item \textbf{Cosine Similarity (cos)}: Embedding similarity between predicted and gold labels.
    \item \textbf{Information Recall ($\text{Recall}_{\text{info}})$}: Proportion of critical diagnostic cues correctly elicited:
    \begin{equation}
    \text{Recall}_{\text{info}} = \frac{|I_T \cap I_{CHA}|}{|I_T|}
    \end{equation}
    \item \textbf{Average Number of Turns ($n$)}: Total dialogue length.
\end{itemize}

\subsection{Experimental Results}

Table~\ref{tab:diagnosis_results} shows DocCHA outperforming all baselines across datasets and metrics. On ICMS, it achieves \textbf{95.86\%} accuracy (+5.18\% over GPT-4o; +29.63\% over LLaMA-3), higher cosine similarity (+3.41), and better cue recall (+2.37). It uses \textbf{7.1} turns on average, showing improved informativeness without unnecessary length.

On DX, DocCHA maintains strong performance (\textbf{94.14\%} accuracy), with consistent semantic and recall gains, demonstrating its robustness across diagnostic domains and languages.
\begin{figure*}[ht]
\centering
\includegraphics[width=1\textwidth]{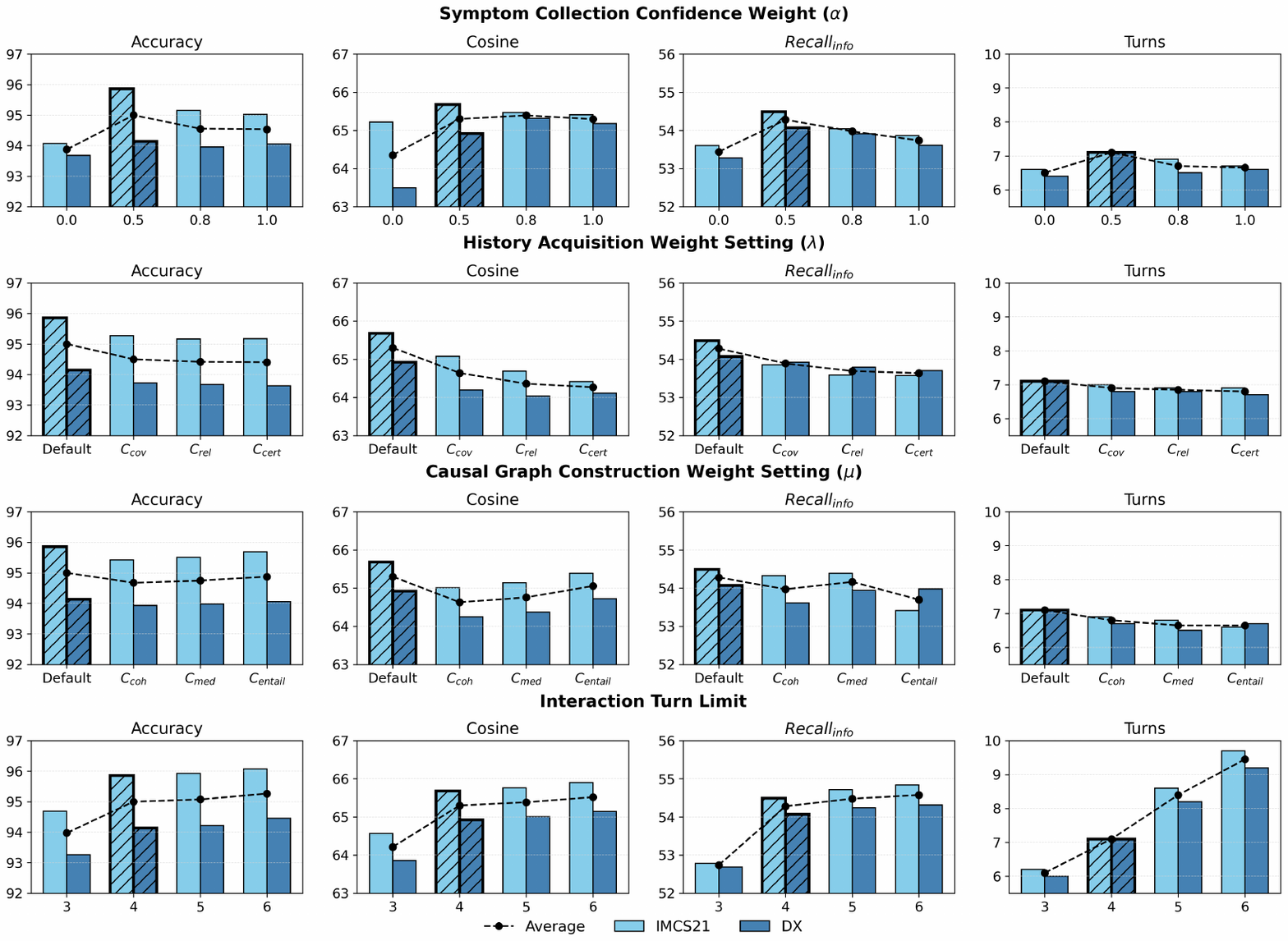}
\caption{Sensitivity of DocCHA’s modules across Accuracy, Cosine Similarity, Information Recall, and Turn Count on IMCS21 and DX. Rows correspond to: Symptom Collection (\boldmath$\alpha$), History Acquisition (\boldmath$\lambda$), Causal Reasoning (\boldmath$\mu$), and Turn Limit. Bold bars indicate the configuration used as the default setting for each module. Black dotted lines represent the average across datasets.}
\label{fig:sensitive}
\end{figure*}
\subsection{Ablation Study}

We ablate each module to measure its contribution:

\paragraph{Module 1 (Symptom Collection).}
Accuracy drops \textbf{2.0--2.4\%}, cosine and recall degrade \textbf{2.3--3.1\%}, and turn count shrinks, indicating under-exploration. This module is critical for extracting discriminative symptoms and ensuring adequate detail for downstream reasoning. Its removal leads to premature decisions with insufficient evidence.

\paragraph{Module 2 (History Confidence).}
Causes \textbf{0.6--0.8\%} accuracy loss and sharper cosine/relevance degradation. The system still asks questions, but with less targeted depth. Without structured confidence scoring, irrelevant or vague history may dilute useful signals, reducing overall clarity of the diagnostic context.

\paragraph{Module 3 (Causal Graph).}
Lowers accuracy by \textbf{1.6\%} with limited effect on turn count, revealing its role in refining final decisions and avoiding hallucinations. The absence of structured reasoning leads to less interpretable and occasionally unsupported conclusions.

\paragraph{Summary.}
All modules are essential, with Module 1 being the most impactful. Joint modeling of structured inquiry and confidence-aware reasoning drives DocCHA’s gains in informativeness and accuracy.

\subsection{Sensitivity Analysis}

We conduct a four-part sensitivity analysis to assess the robustness of DocCHA's confidence-weighted design, focusing on the weighting schemes for each module ($\alpha$, $\lambda$, $\mu$) and the turn limit $T$. Results (Figure~\ref{fig:sensitive}) are consistent across both datasets and show the system's resilience under varied parameter settings.

\textbf{Symptom Collection ($\alpha$):} Performance peaks at $\alpha{=}0.5$, balancing coverage and detail completeness. Overemphasizing either dimension harms recall and reasoning accuracy, confirming that ambiguity reduction requires both breadth and specificity.

\textbf{History Acquisition ($\lambda$):} The default weighting (coverage=0.4, relevance=0.3, certainty=0.3) yields optimal results. Prioritizing only fluency or confidence weakens performance, underscoring the importance of broad factual grounding in patient history.

\textbf{Causal Reasoning ($\mu$):}
Giving coherence the highest weight ($\mu_1{=}0.4$) produces the most accurate and interpretable reasoning. Over-reliance on ontology grounding degrades quality, highlighting that medical knowledge supports—but does not replace—logical fluency.

\textbf{Turn Limit ($T$):} Increasing turns from 3 to 5 significantly improves performance; beyond that, gains plateau. This validates DocCHA’s adaptive stopping: sufficient information can be gathered early, avoiding unnecessary user burden.

\section{Conclusion}

We introduced \textbf{DocCHA}, a confidence-aware, modular diagnostic agent that emulates real-world clinical reasoning through structured multi-turn dialogue. By integrating discriminative symptom collection, multi-dimensional history acquisition, and causal graph-based reasoning, DocCHA significantly improves diagnostic accuracy and information recall over strong LLM baselines—while maintaining interaction efficiency. Our experiments and sensitivity analysis validate the robustness and interpretability of this design, showcasing its potential for building reliable and adaptive LLM-powered health agents. This framework also lays a foundation for broader deployment in low-resource or multilingual clinical settings, where trust, transparency, and diagnostic completeness are essential.

\bibliography{acl_latex}

\appendix
\section{Appendix: Prompt Design Details}
\label{sec:appendix}

To support transparency and reproducibility, we provide the full prompting strategies used by each diagnostic module in \textbf{DocCHA}. Each prompt is carefully aligned with its corresponding confidence metric (e.g., coverage, relevance, certainty), and designed to operate under a multi-turn dialogue paradigm. The prompts follow a chain-of-thought structure that (1) extracts current facts, (2) infers probable diagnoses, and (3) generates next-step questions or reasoning outputs. All prompts were manually designed and tested to ensure clarity, controllability, and generalizability.

\vspace{1mm}
\paragraph{Module 1: Symptom Collection.}
\begin{figure}[ht]
    \centering
    \includegraphics[width=\linewidth]{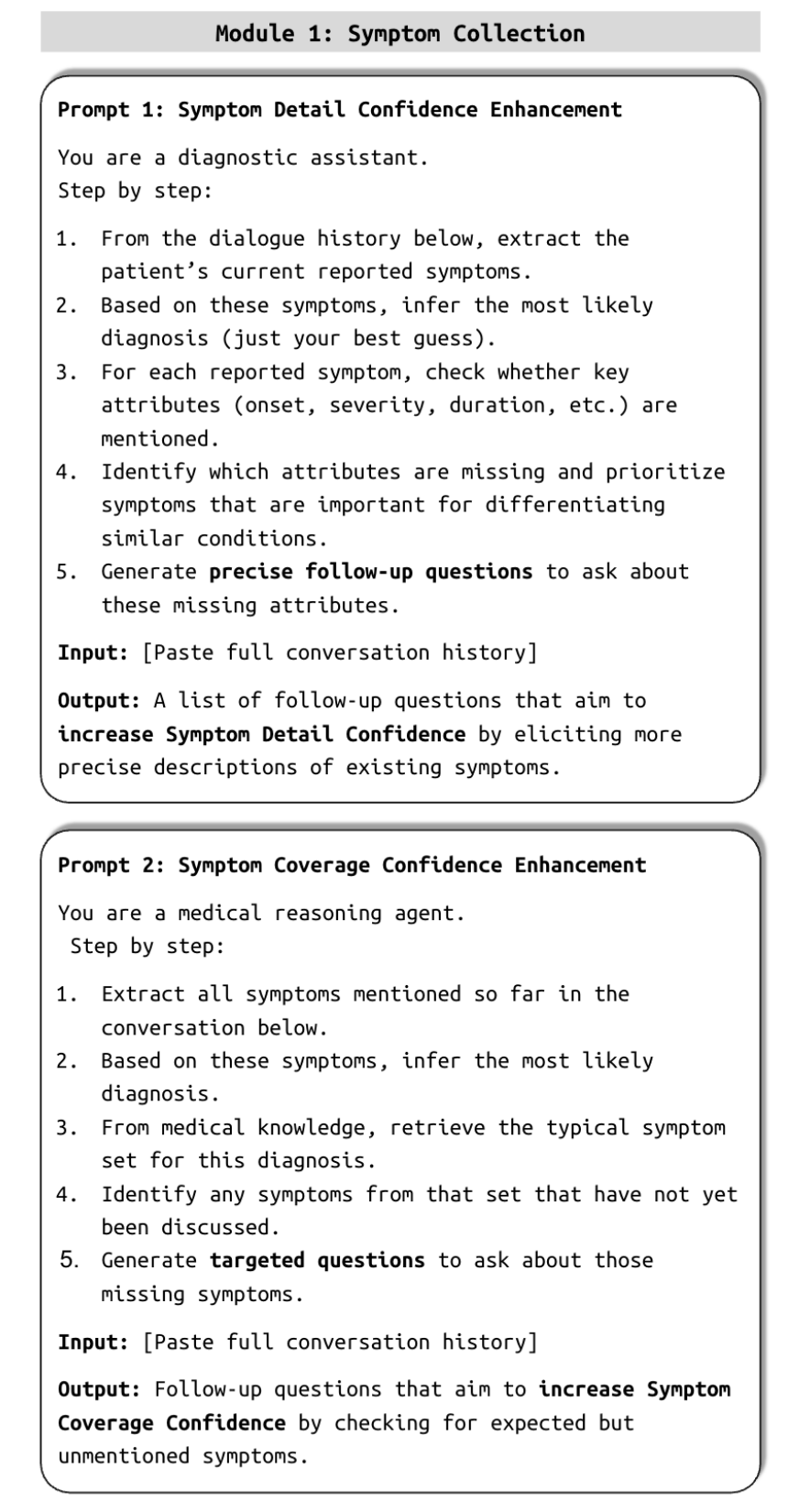}
    \caption{Prompt instructions used in Module 1 for increasing \textit{Symptom Detail Confidence} (Prompt 1) and \textit{Symptom Coverage Confidence} (Prompt 2).}
    \label{fig:module1_prompt}
\end{figure}

As shown in Figure~\ref{fig:module1_prompt}, Module 1 uses two distinct prompts depending on the specific deficiency detected in the current symptom profile:

\begin{itemize}[leftmargin=*,itemsep=0pt]
    \item \textbf{Prompt 1 – Detail Completion}: Given a set of user-reported symptoms, the LLM is instructed to identify missing attributes such as onset time, severity, and duration. The prompt includes examples of well-documented vs. under-specified symptoms to encourage detailed follow-up questions.
    
    \item \textbf{Prompt 2 – Symptom Coverage Expansion}: The LLM is asked to infer a probable diagnosis based on the current symptom set, then compare that diagnosis against expected symptom manifestations. Missing but diagnostically discriminative symptoms are identified, and appropriate follow-up questions are generated.
\end{itemize}

These prompts support dynamic elicitation and improve both the breadth and contextual richness of symptom information.

\vspace{1mm}
\paragraph{Module 2: History Acquisition.}
\begin{figure}[ht]
    \centering
    \includegraphics[width=\linewidth]{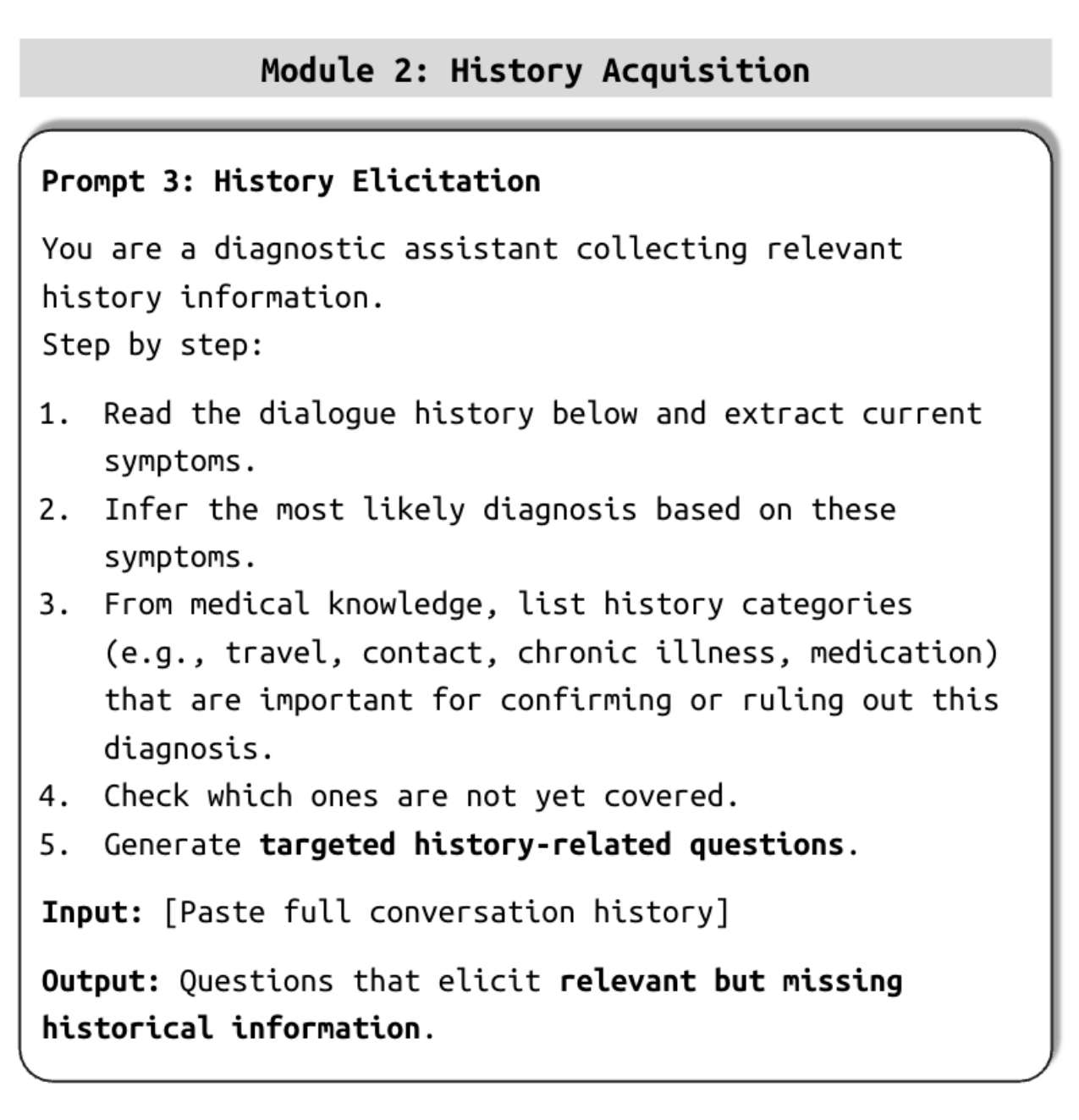}
    \caption{Prompt used in Module 2 to elicit missing contextual history relevant to the most likely diagnosis.}
    \label{fig:module2_prompt}
\end{figure}

Figure~\ref{fig:module2_prompt} presents the prompt used for history elicitation. After observing the current symptoms and preliminary diagnosis, the LLM is instructed to identify high-priority contextual history categories (e.g., medication use, exposure, chronic illness, family history) that are typically relevant to the inferred diagnosis. For each missing or low-confidence category, the LLM generates a specific, non-redundant follow-up question.

In addition, the prompt includes soft instructions for disambiguating vague user history (e.g., ``maybe'' or ``occasionally'') and encouraging more certain statements. This supports improvement in both \textit{coverage} and \textit{certainty} dimensions of the history confidence score.

\vspace{1mm}
\paragraph{Module 3: Causal Graph Construction and Refinement.}
\begin{figure}[ht]
    \centering
    \includegraphics[width=\linewidth]{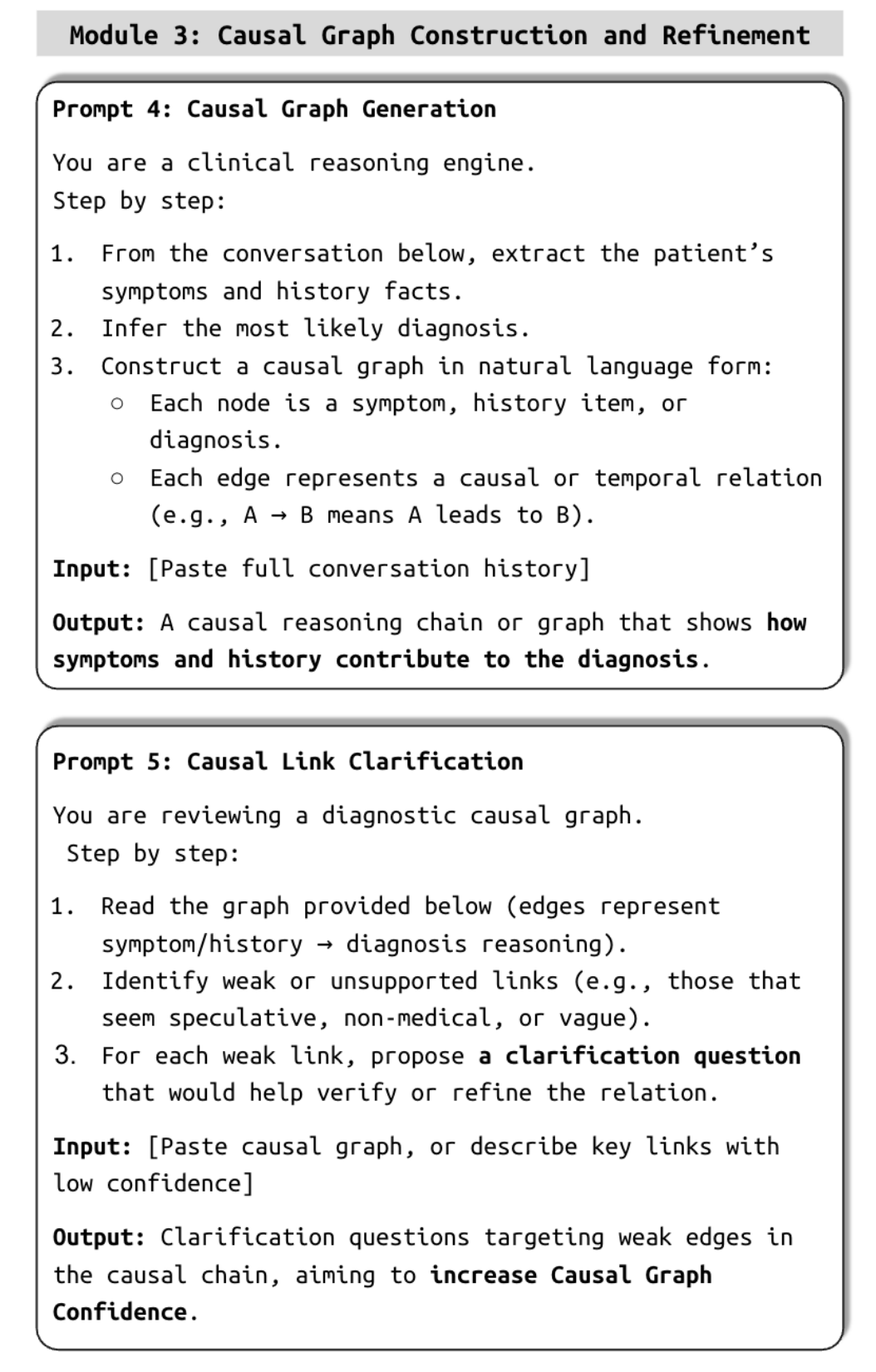}
    \caption{Prompt set used in Module 3 for generating causal graphs (Prompt 4) and refining weak reasoning links (Prompt 5).}
    \label{fig:module3_prompt}
\end{figure}

Figure~\ref{fig:module3_prompt} includes two prompts used for causal reasoning:

\begin{itemize}[leftmargin=*,itemsep=0pt]
    \item \textbf{Prompt 4 – Causal Graph Generation}: Given structured inputs (symptoms and history), the LLM is asked to construct a stepwise, natural-language explanation chain linking the evidence to a diagnostic hypothesis. The output is scored on coherence, medical plausibility (based on UMLS alignment), and entailment confidence.
    
    \item \textbf{Prompt 5 – Weak Link Clarification}: When causal links are flagged as uncertain or unsupported, this prompt guides the LLM to explicitly identify which part of the reasoning is weakest, and suggest a clarification question targeting that link. This enables targeted reasoning repair without regenerating the entire explanation.
\end{itemize}

Together, these prompts enable interpretable and modular diagnostic reasoning, and support confidence-aware refinement based on weak signal identification.

\vspace{1mm}
\paragraph{Implementation Notes.}
All prompts are implemented as single-turn calls to the LLM, using chain-of-thought reasoning and contextual memory of previous dialogue states. Prompts are parameterized with candidate diagnoses, observed symptom/historical facts, and module-specific confidence outputs. The LLM used is either GPT-4o or LLaMA-3 depending on the experiment configuration.

This modular and interpretable prompting framework is central to DocCHA’s ability to perform structured, adaptive diagnostic dialogue under realistic interaction constraints.

\end{document}

%% file: tables/evaluation_res.tex
\begin{table*}[ht]
\centering
\caption{Diagnostic Accuracy and Module Contribution Analysis on IMCS21 and DX Datasets}
\label{tab:diagnosis_results}
\small
\begin{tabular}{lcccccccc}
\toprule
\multirow{2}{*}{\textbf{Method}} & \multicolumn{4}{c}{\textbf{ICMS Dataset}} & \multicolumn{4}{c}{\textbf{DX Dataset}}\\
\cmidrule(lr){2-5} \cmidrule(lr){6-9}
 & Acc. & $\cos$ & $\text{Recall}_{\text{info}}$ & $n$ & Acc. & $\cos$
 & $\text{Recall}_{\text{info}}$ & $n$ \\
\midrule

\rowcolor{gray!10}
\multicolumn{9}{l}{\textbf{LLMs with Direct Prompting}} \\
LLaMA-3   & 66.23 & 56.07  & 47.72 & 5.6 & 65.72 & 55.48 & 46.90 & 5.1\\
GPT-3.5   & 87.69 & 62.20  & 52.04 & 6.4 & 85.44 & 62.03 & 51.75 & 5.9 \\
GPT-4o    & 90.68 & 62.27  & 52.12 & 6.8 & 89.96 & 62.23 & 51.49 & 6.4 \\
\rowcolor{gray!10}
\multicolumn{9}{l}{\textbf{DocCHA Model Series}} \\
DocCHA (GPT-4o)  & \textbf{95.86} & \textbf{65.68}  & \textbf{54.49} & \textbf{7.1} & \textbf{94.14} & \textbf{64.92} & \textbf{54.07} & \textbf{7.1} \\
\hspace{1em} w/o Module 1  & 
93.56 \scriptsize{(-2.4\%)} & 
64.18 \scriptsize{(-2.3\%)} & 
53.09 \scriptsize{(-2.6\%)} & 
6.2 & 
92.23 \scriptsize{(-2.0\%)} & 
62.91 \scriptsize{(-3.1\%)} & 
52.86 \scriptsize{(-2.2\%)} & 
6.1 \\

\hspace{1em} w/o Module 2  & 
95.07 \scriptsize{(-0.8\%)} & 
63.85 \scriptsize{(-2.8\%)} & 
54.27 \scriptsize{(-0.4\%)} & 
6.8  & 
93.56 \scriptsize{(-0.6\%)} & 
63.83 \scriptsize{(-1.7\%)} & 
53.36 \scriptsize{(-1.3\%)} & 
6.7 \\

\hspace{1em} w/o Module 3  & 
94.29 \scriptsize{(-1.6\%)} & 
63.62 \scriptsize{(-3.1\%)} & 
54.06 \scriptsize{(-0.8\%)} & 
6.6 & 
92.67 \scriptsize{(-1.6\%)} & 
63.35 \scriptsize{(-2.4\%)} & 
53.17 \scriptsize{(-1.7\%)} & 
6.5 \\

\bottomrule
\end{tabular}
\end{table*}